\documentclass{article}

\usepackage{microtype}
\usepackage{graphicx}
\usepackage{subcaption}
\usepackage{booktabs}

\usepackage{hyperref}

\usepackage[accepted]{arxiv}

\usepackage{amsmath}
\usepackage{amssymb}
\usepackage{mathtools}
\usepackage{amsthm}
\usepackage{caption}
\usepackage{listings}
\usepackage{xcolor}
\usepackage{multirow}

\usepackage[capitalize,noabbrev]{cleveref}

\theoremstyle{plain}

\theoremstyle{definition}

\theoremstyle{remark}

\usepackage[textsize=tiny]{todonotes}

\renewcommand{\paragraph}[1]{\vspace{-0.1em} \noindent \textbf{#1}}
\newcommand{\code}[1]{\texttt{\small{#1}}}

\icmltitlerunning{QMoE: Practical Sub-1-Bit Compression of Trillion-Parameter Models}

\begin{document}

\twocolumn[
\icmltitle{QMoE: Practical Sub-1-Bit Compression of Trillion-Parameter Models}

\icmlsetsymbol{equal}{*}

\begin{icmlauthorlist}
    \icmlauthor{Elias Frantar}{ista}
    \icmlauthor{Dan Alistarh}{ista,nm}
\end{icmlauthorlist}

\icmlaffiliation{ista}{Institute of Science and Technology Austria (ISTA)}
\icmlaffiliation{nm}{Neural Magic Inc}

\icmlcorrespondingauthor{Elias Frantar}{elias.frantar@ist.ac.at}

\vskip 0.3in
]

\printAffiliationsAndNotice{}

\begin{abstract}
Mixture-of-Experts (MoE) architectures offer a general solution to the high inference costs of large language models (LLMs) via sparse routing, bringing faster and more accurate models, at the cost of massive parameter counts. For example, the SwitchTransformer-c2048 model
has 1.6 trillion parameters, requiring 3.2TB of accelerator memory to run efficiently, which makes practical deployment challenging and expensive. In this paper, we present a solution to this memory problem, in form of a new compression and execution framework called QMoE. Specifically, QMoE consists of a scalable algorithm which accurately compresses trillion-parameter MoEs to less than 1 bit per parameter, in a custom format co-designed with bespoke GPU decoding kernels to facilitate efficient end-to-end compressed inference, with minor runtime overheads relative to uncompressed execution. Concretely, QMoE can compress the 1.6 trillion parameter SwitchTransformer-c2048 model to less than 160GB (20x compression, 0.8 bits per parameter) at only minor accuracy loss, in less than a day on a single GPU. This enables, for the first time, the execution of a trillion-parameter model on affordable commodity hardware, like a single server with 4x NVIDIA A6000 or 8x NVIDIA 3090 GPUs, at less than 5\% runtime overhead relative to ideal uncompressed inference. The source code and compressed models are available at \url{github.com/IST-DASLab/qmoe}.
\end{abstract}

\section{Introduction}

Generative large language models (LLMs), e.g.~\cite{radford2019language, brown2020language, touvron2023llama, touvron2023llama2}, have garnered significant industrial and popular attention due to their surprising performance across many practical language and reasoning tasks. 
Yet, a major obstacle to broad deployment is given by their extremely high inference costs.
One particularly promising approach for reducing these costs is the use of \emph{Mixture-of-Experts (MoE)} architectures, e.g.~\cite{fedus2022switch, artetxe2021efficient}, 
whose general idea is to replicate certain model components many times while routing each input \emph{only to a small subset of those replicas}.
Through expert ``specialization'' to input subsets, MoEs achieve faster inference for the same model quality, but with significantly higher memory costs due to components being replicated hundreds or even thousands of times, for the largest and best-performing models.

For example, the popular SwitchTransformer family~\citep{fedus2022switch}, which we focus on in this study, uses between 128 and 2048 experts (layer replicas) to significantly outperform standard dense T5 models~\citep{raffel2020exploring} in terms of inference and training costs, at equivalent model accuracy. \citet{artetxe2021efficient} report similar improvements, on different tasks, for 512 experts. However, these results come at the cost of dramatic increases in model size: the largest SwitchTransformer has 1.6 trillion parameters, requiring 3.2TB of storage in standard half-precision, and correspondingly requires a hundred or more expensive (GPU or TPU) accelerators for efficient usage. This not only makes practical deployment costly and challenging, but also strongly limits research on such models.

\paragraph{Challenges.} It is natural to ask whether the truly massive memory costs of such MoEs can be reduced via standard techniques for \emph{model compression}, such as quantization~\cite{gholami2021survey} or sparsity~\cite{hoefler2021sparsity}, without significant accuracy loss. Achieving this would require overcoming conceptual and technical barriers:

\begin{enumerate}
    \itemsep0pt
    \item Conceptually, existing post-training compression methods, whose costs would be affordable enough to execute on such models, are currently only able to reduce precision to 3 or 4 bits per parameter \citep{frantar2022gptq, dettmers2022case, wu2023zeroquant} or around 50\% sparsity \citep{frantar2023sparsegpt}, before significant accuracy loss occurs. Yet, making trillion-parameter MoEs practical would require compression rates between $10\times$ and $20\times$ relative to 16-bit precision, i.e., on average \emph{less than 1 bit per parameter}.
    \item A key practical issue is \emph{scaling}: applying state-of-the-art compression methods, designed for large dense models, to MoEs that are an order of magnitude larger, while maintaining affordability, runs into a plethora of memory, performance and reliability roadblocks.
    \item Actually achieving \emph{sub-1-bit} compression would require a non-trivial custom compression format. Such a format would also need to come with decoding algorithms that are highly-efficient on accelerators such as GPUs, in order to run inference on compressed models without major processing slowdowns.
\end{enumerate}

\paragraph{Contribution.} In this paper, we overcome these challenges, and introduce QMoE, a framework for accurate compression and fast compressed inference of massive MoEs, reducing model sizes by 10--20$\times$, to less than 1 bit per parameter. QMoE is specifically designed to compress and subsequently inference with models like the 1.6 trillion parameter SwitchTransformer-c2048, using only modest computational resources.

Our key technical contributions are a highly scalable compression algorithm implementation and a customized compression format designed together with bespoke GPU-kernels for fast on-the-fly decoding. Further, we show for the first time that accurate sub-1-bit compression of trillion parameter MoEs is feasible and can be achieved via affordable retraining-free compression techniques.

Concretely, we reduce the size of SwitchTransformer-c2048, the largest openly-available model, from 3.2TB in bfloat16 to less than 160GB in our customized compressed format, that is, $\approx 0.8$ bits per parameter, at only a minor increase in loss on pretraining validation and zero-shot data. Using our QMoE kernels, this compressed model can then be executed fully, without any slow offloading, on commodity hardware such as $8\times$ NVIDIA RTX 3090 or $4\times$ NVIDIA A6000 GPUs, with $< 5\%$ runtime overhead relative to an idealized version of uncompressed execution, which would require $\approx 20\times$ more GPUs.

In summary, our work enables, for the first time, the performant execution of massive-scale MoE models on commodity hardware. This is illustrated by the fact that we are able to efficiently run the trillion-parameter SwitchTransformer-c2048 model on a single commodity GPU server, with minor accuracy loss. This addresses one of the key limitations behind MoE architectures, and should improve their practical adoption as well as facilitate further research on understanding and improving such models.

\section{Background}

\subsection{Mixture of Expert Models (MoEs)}

\begin{figure}[ht!]
    \centering
    \includegraphics[width=.9\linewidth]{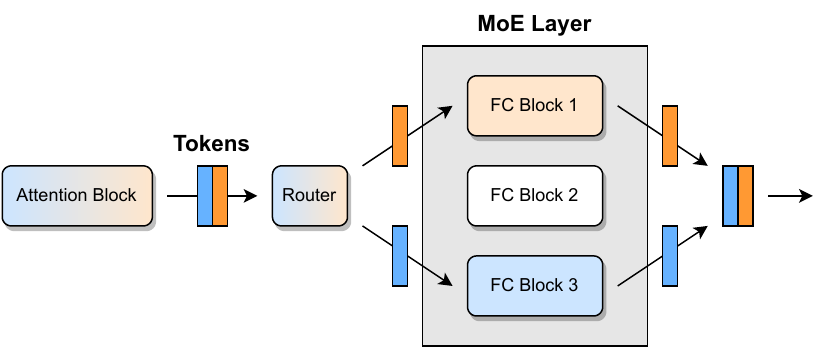}
    \vspace{-5pt}
    \caption{Example of an MoE Transformer block. Each token is routed to a different fully-connected (FC) block.}
    \label{fig:moe-example}
\end{figure}

The core idea behind Mixture of Expert models (MoEs) is to increase the number of parameters, and thus the network's modelling power, while at the same time keeping compute costs near-constant, relative to a standard feed-forward architecture. This is typically achieved by creating many copies of certain model components, each of which is responsible for processing only a subset of all input tokens. The corresponding input-to-component assignments are generally decided by a ``router'' layer. Probably the most common MoE design \citep{fedus2022switch, artetxe2021efficient}, which we also focus on in this paper, is to replicate the fully-connected module of a Transformer and route tokens to the replica, referred to as an \textit{expert}, with the highest assignment score predicted by a linear routing layer; see Figure~\ref{fig:moe-example} for an illustration. This design enables efficient training and inference of extremely large models, using 100s or even 1000s of experts/, since each token is processed only by a small subset of the massive overall network.

MoEs have been shown to bring substantial accuracy and training speed improvements for equivalent inference speed \citep{clark2022unified, du2022glam, zoph2022st}. However, their current practicality is limited since they are extremely large in size and thus require massive amounts of accelerator memory to be executed efficiently.

\subsection{Data-dependent Quantization}

The currently most effective strategy for reducing model size and corresponding memory costs is \textit{quantization}, i.e., converting model weights to lower numerical precision. On large models \citep{dettmers2022llm, dettmers2022case}, in particular also MoEs \citep{kim2022says, yi2023edgemoe}, just simple rounding can decrease precision to 8 or even 4 bits per weight, at minimal accuracy loss relative to the standard half (16-bit) precision employed for these models. However, some MoEs are so large that reduction rates significantly higher than $4\times$ (accomplished by 4-bit) would be required to render them practical. Accurately quantizing models to extremely low precision (e.g., lower than 3 bits per parameter) typically requires more sophisticated \textit{data-dependent} methods \cite{nagel2020up, wang2020towards, hubara2021accurate}.

Such data-dependent quantization methods use a small set of calibration data, which is passed through the model. As this happens, for each linear layer $\ell$ with weights $W_
{\ell}$, quantized weights $Q_\ell$ are determined one-by-one. Specifically, one approach to do this is by solving a layer-wise quantization problem, stated with respect to $W_\ell$ and the observed calibration data inputs $X_\ell$ at the current layer:
\begin{equation}
    \label{eq:layerwise}
    \text{argmin}_{Q_\ell} \, ||Q_\ell X_\ell - W_\ell X_\ell||.
\end{equation}
Various solvers for Equation~(\ref{eq:layerwise}) have been proposed, with some optimized, in terms of speed and accuracy, particularly for extremely large models, like GPTQ \citep{frantar2022gptq} or ZeroQuant \citep{yao2022zeroquant, wu2023zeroquant}. The former performs quantization using second-order information in the layer-wise Hessian matrix $X_\ell X_\ell^\top$, while the latter applies SGD-optimization with straight-through gradient estimation \citep{bengio2013estimating}.

Another noteworthy characteristic of many such methods is that per-layer quantization can be performed \textit{sequentially}, using the input from the already partially quantized model up to layer $\ell - 1$, when quantizing layer $\ell$, serving to reduce error accumulation. Concretely, this can be efficiently implemented by using $X_\ell$ to find $Q_\ell$ before passing on $X_{\ell + 1} = Q_\ell X_\ell$ to the next layer.

\subsection{MoE Quantization}

There are several aspects which make very-low-bit, e.g. ternary (3 values) quantization promising for MoE models:

\begin{itemize}
    \itemsep0pt
    \item In many architectures, almost all parameters are located in the experts, as they are 1000s of them. This means that, for size reduction, it suffices to focus on compressing just those experts and leave other layers in standard precision. This reduces error accumulation since only a subset of modules involved in a forward pass are actually quantized.
    \item Previous work has observed that extremely large dense models are more resistant to quantization noise than smaller ones \citep{frantar2022gptq, chee2023quip}. Large MoEs can be much larger than some of these massive dense models, and are thus a prime target for accurate quantization.
    \item MoE training involves additional stochasticity through routing instabilities and strategies like token dropping~\citep{lepikhin2020scaling}, which may inherently encourage high resistance to noise. Finetuning is also often performed with high dropout \citep{fedus2022switch}.
\end{itemize}

Our experiments in Section~\ref{sec:compression-results} confirm that MoEs are indeed highly robust to extreme levels of quantization.

\section{Scaling Data-dependent Quantization to Trillion Parameter MoEs}
\label{sec:scaling-quantization}

\subsection{Challenges}
\label{sec:challenges}

While data-dependent quantization techniques have already been used to successfully compress large dense models up to 176 billion parameters \citep{frantar2022gptq, wu2023zeroquant}, applying them to \textit{sparse mixture-of-expert models another order of magnitude larger} brings several new challenges.

\paragraph{Memory Costs.}
The first major problem we encounter is a large increase in the memory required to apply such techniques. Not only are the original model weights nearly $10\times$ larger, but the quantization process itself also needs $> 100\times$ more data. The latter constraint is because accurate data-dependent quantization methods require a sufficient number of input samples for each layer that is being compressed. For very large dense models, a few hundreds of thousands of ``calibration tokens'' typically suffice \citep{frantar2022gptq, yao2022zeroquant}. However, in MoEs with thousands of layers, a single expert processes only a small subset of all inputs, hence we need much more tokens overall to achieve good coverage of all experts. Further, in encoder-decoder architecture models, like SwitchTransformers, each token is processed only by half of the model, again increasing data requirements. For \emph{fast} compression, we must maintain intermediate results for the full calibration dataset, which requires 100s of GBs of memory for the largest models.

\paragraph{GPU Utilization.}
The next significant challenge is that existing large-scale quantization implementations, in particular for GPTQ and related methods \citep{frantar2022gptq, chee2023quip}, are designed to be fast and memory efficient for the massive individual layers occurring in dense models. Meanwhile, MoEs typically have smaller layers, but $100\times$ to $1000\times$ more of them. Current implementations have poor GPU utilization in this case, and consequently bad performance. A similar issue occurs if activations and weights have to be transferred between CPU and GPU with high frequency, which may be required to cope with the massive memory requirements discussed previously.

\paragraph{Reliability Requirements.}
Finally, another issue when compressing models with tens of thousands of layers is that running into rare edge cases, which may break the process, is highly likely. This is includes numerical problems like non-invertible layer-wise Hessians, as well as model-specific ones, e.g., extreme routing patterns on particular layers.

\subsection{System Design \& Optimizations}
\label{sec:system-optimizations}

In this section, we describe system-level design and optimizations to address the challenges in Section~\ref{sec:challenges}. This allows us to apply data-dependent compression to massive MoEs, while preserving the key feature of post-training compression techniques: the ability to perform effective compression using only modest computational resources, e.g., a single NVIDIA A6000 GPU and less than one day of compute. Although we focus on scaling the popular GPTQ method, most techniques described below will generalize to other data-dependent quantization approaches, like ZeroQuant \citep{yao2022zeroquant}, as well.

\subsubsection{Optimized Activation Offloading}
\label{sec:offloading}

\begin{figure*}
    \centering
    \includegraphics[width=.75\textwidth]{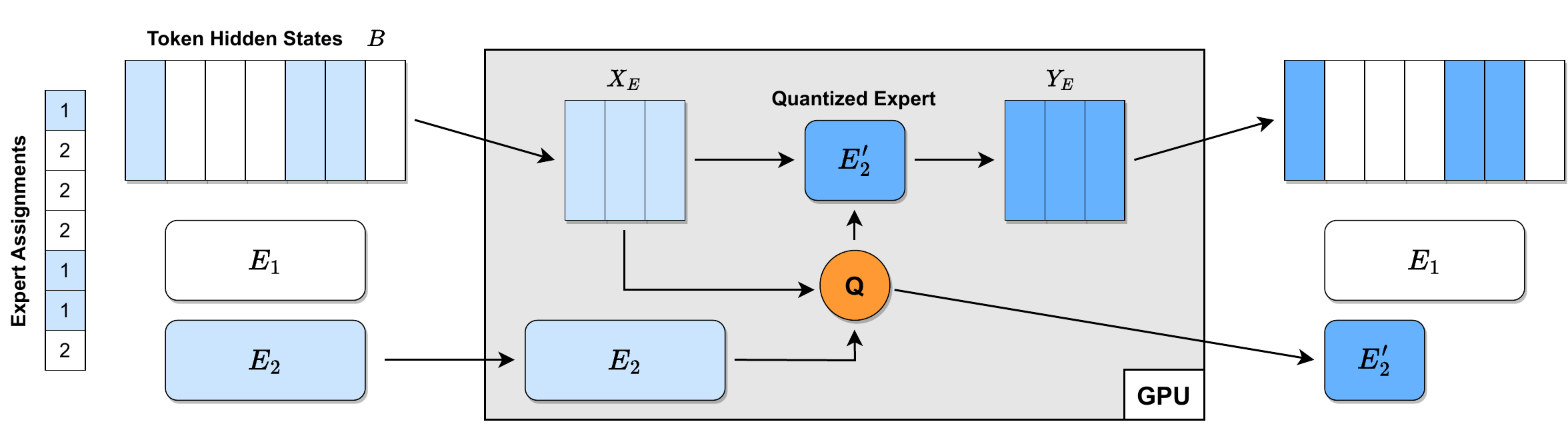}
    \vspace{-5pt}
    \caption{Illustration of the offloading execution for the sparse part of a Transformer block. An expert $E_2$ and its corresponding input tokens $X_E$ are fetched to GPU memory to produce $E'_2$, which together with the corresponding outputs $Y_E$ are written back to CPU again.}
    \label{fig:offloading}
\end{figure*}

As discussed in Section~\ref{sec:challenges}, a key challenge in compressing MoEs is that we need to maintain massive activation sets. Yet, it is possible to carefully orchestrate model execution in such a way that we only ever need to perform computation on a small subset of the intermediate data. This allows us to offload main storage from GPU, to much less expensive and plentiful CPU memory.

Concretely, we maintain a single large buffer $B$ which we update as follows, for the dense part of a Transformer block:

\begin{enumerate}
    \itemsep0pt
    \item Fetch one ``sample'' $X$, containing a few hundreds of tokens, from CPU to GPU.
    \item Pass it through the corresponding dense layers to obtain the result $Y$.
    \item Calculate and store expert assignment for tokens in $Y$.
    \item Send $Y$ back to CPU and overwrite $X$ in $B$.
\end{enumerate}

and respectively for the sparse part, looping over experts:

\begin{enumerate}
    \itemsep0pt
    \item Fetch all individual tokens in $B$ that have been assigned to expert $E$, denoted by $X_E$, from CPU to GPU.
    \item Use them to produce compressed expert $E'$ (for example, with GPTQ).
    \item Run $X_E$ through $E'$ to get $Y_{E'}$.
    \item Send $Y_{E'}$ back to CPU and overwrite $X_E$ in $B$.
\end{enumerate}

This process, which is visualized in Figure~\ref{fig:offloading}, minimizes both memory consumption and transfer cost: we need only a single copy of $B$ and each token is only read and written twice per Transformer block.

\begin{figure}[ht!]
    \centering
    \includegraphics[width=.85\linewidth]{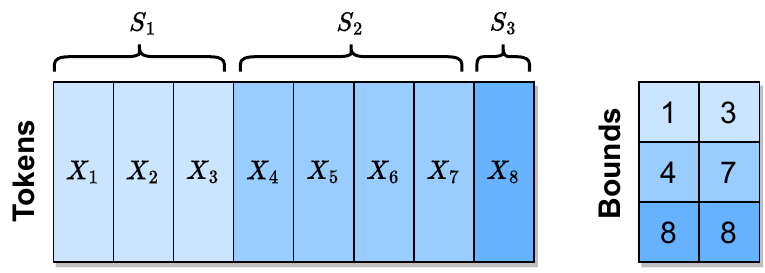}
    \vspace{-5pt}
    \caption{List buffer example with 3 samples, indicated by hue.}
    \label{fig:list-buffer}
\end{figure}

\subsubsection{List Buffer}

To efficiently support per-sample access for evaluating dense model components, as well as fully-vectorized querying of expert tokens, we store $B$ as a \textit{list buffer} data structure. This can be seen as a huge contiguous buffer of all token hidden states, together with delimiter indices denoting boundaries between individual samples. Figure~\ref{fig:list-buffer} illustrates this storage format. This datastructure is crucial for efficiency; naively iterating over samples and fetching relevant tokens via masking is unusably slow for large sample counts.

\subsubsection{Lazy Weight Fetching}

Since the weights of the 1.6 trillion parameter model consume $> 3$ TB of storage, they cannot even be stored in CPU RAM. Thus, we lazily fetch them directly from disk storage as they are required. If we follow the inference procedure outlined in Section~\ref{sec:offloading}, this would be exactly once. Afterwards, their memory is released again.

\subsubsection{Expert Grouping}

Additionally, in order to avoid GPU underutilization (see Section \ref{sec:challenges}), we group multiple experts together and apply a joint \emph{batched variant} of the GPTQ algorithm. Concretely, we extract the inputs $X_E$ corresponding to all experts $E \in \mathcal{E}$ in group $\mathcal{E}$ (the $X_E$ will generally have different sizes) and compute Hessians $H_E$. These matrices, together with the weight matrices $W_E$, are then stacked to 3-dimensional tensors, on which our modified GPTQ algorithm operates, compressing all experts simultaneously. We can also compute $H_E = X_E X_E^\top$ directly with a single matmul as the $X_E$ are generally small enough, avoiding the slow per-sample accumulation employed by prior implementations. Our default expert groupsize $|\mathcal{E}|$ is 16, which brings a good trade-off between GPU memory consumption and utilization.

Table~\ref{tab:gpu-util} demonstrates the impact of expert grouping via GPTQ batching, when compressing a sparse encoder layer of switch-base-128 using 10k samples; $|\mathcal{E}| = 16$ yields about $\approx 6\times$ speedup over standard per-expert computation. 

\begin{table}[ht!]
    \centering
    \begin{tabular}{|ccc|}
        \toprule
        $|\mathcal{E}| = 1$ & $|\mathcal{E}| = 4$ & $|\mathcal{E}| = 16$ \\
        \midrule
        174.1s & 54.4s & \textbf{28.8s} \\
        \bottomrule
    \end{tabular}
    \vspace{-5pt}
    \caption{Sparse layer compression time for different $|\mathcal{E}|$.}
    \label{tab:gpu-util}
\end{table}

\subsubsection{Robustness Modifications}

To achieve sufficiently high robustness for successfully quantizing trillion parameter models with tens of thousands of layers, we need to employ various numerical and memory adjustments. The most important are listed below:

\begin{itemize}
    \itemsep0pt
    \item We use $10\times$ higher relative Hessian dampening $\delta = 0.1$, avoiding breakdowns with inf-values.
    \item Very few layer Hessians are not invertible even after high dampening; we skip GPTQ for those and simply perform vanilla rounding.
    \item Sometimes an expert receives a number of tokens that is much larger than average, leading to out-of-memory situations when these are fetched to GPU. We avoid this by capping the maximum number of tokens used for compression at $4\times$ the mean and use multiple iterations for computing and updating $Y_E$ in such cases.
\end{itemize}

\subsection{Accuracy Improvements}

In addition to implementing a highly efficient compression system, we also make new discoveries about applying GPTQ in our particular context, i.e., for models trained for masked-language-modelling, MoEs and ternary quantization.

\paragraph{Premasking Special Tokens.}
First, we find that results can be improved if the various special separator tokens inserted by the masked-language-modelling task \citep{raffel2020exploring} are excluded from the calibration data used for compression. Conretely, in the encoder, we mask out those ``mask-tokens'' during the Hessian computation. Meanwhile, in the decoder, we skip the token directly \textit{before} such a special token as this is the one used to predict the latter.

As shown in Table~\ref{tab:masking} for switch-base-128 with 10k samples, this brings noticeably lower loss at no additional compute cost. We think that because those tokens are very common during training, the model is so robust in their prediction that any error compensation on them during quantization is unnecessary, while worsening correction for other tokens.

\begin{table}[ht!]
    \centering
    \begin{tabular}{|c|ccc|}
        \toprule
        mask & BF16 & 2bit & tern \\
        \midrule
        no & 1.73 & 1.86 & 2.16 \\
        yes & 1.73 & \textbf{1.76} & \textbf{1.99} \\
        \bottomrule
    \end{tabular}
    \vspace{-5pt}
    \caption{Impact of special token masking; validation loss.}
    \label{tab:masking}
\end{table}

\paragraph{Ineffective Heuristics.}
We also evaluate two more recently proposed GPTQ enhancement heuristics: activation reordering and true sequential execution \citep{gptq-repo}. However, as shown in Table~\ref{tab:heuristics} for ternary quantization of switch-base-128, we find the former to be actually harmful and the latter to be more or less quality neutral, for our particular use-case. We suspect that, in this highly aggressive setting, quantizing all the most sensitive columns first, leads to large changes of the entire weight matrix, and thus to  overfitting.

\begin{table}[ht!]
    \centering
    \begin{tabular}{|c|ccc|}
        \toprule
        GPTQ & act & seq & act + seq \\
        \midrule
        \textbf{1.99} & 2.23 & \textbf{1.99} & 2.28 \\
        \bottomrule
    \end{tabular}
    \vspace{-5pt}
    \caption{Activation reordering (act) and sequential execution (seq).}
    \label{tab:heuristics}
\end{table}

\section{Realizing Sub-1-Bit Compression}
\label{sec:realizing-sub1}

Using our system discussed in Section~\ref{sec:scaling-quantization}, we can accurately quantize extremely large SwitchTransformers to very low bit-widths: 2-bit and even ternary (3 possible values).
Yet, in practice, this falls still short of our compression goal of less than 1 bit per parameter. 
We find that compression rates can be pushed significantly further by taking advantage of the \emph{low entropy in the quantized weights}. Next, we co-design an encoding scheme and a CUDA kernel which realize sub-1-bit per weight compression in practice, at minimal cost in terms of GPU execution overhead for inference.

\subsection{Natural Sparsity}

We pick quantization grids in standard fashion: row-wise around the min and max weights values \citep{dettmers2022llm, frantar2022gptq}, e.g., for ternary: $\{w_\text{min}, 0, w_\text{max}\}$. These rather wide grids combined with the fact that weights are typically close to normally distributed, \textit{naturally} lead to high sparsity after quantization, i.e., a large number of zeros. We demonstrate this in Table~\ref{tab:sparsity}, averaged over all layers. For ternary weights, the largest model achieves close to \emph{90\% natural sparsity}; the standard deviation is also quite low, at $<$ 5\%. Seen another way, the quantized weights have low entropy, meaning that, on average, significantly less bits per weight should be required for lossless storage.

\begin{table}[!ht]
    \centering
    \begin{tabular}{|c|cc|}
        \toprule
        model & 2-bit & ternary \\
        \midrule
        base128 & 72.2\% & 85.7\% \\
        large128 & 73.1\% & 86.4\% \\
        c2048 & 76.5\% & 88.6\% \\
        \bottomrule
    \end{tabular}
    \vspace{-5pt}
    \caption{Natural sparsity for different compressed models.}
    \label{tab:sparsity}
\end{table}

\subsection{From Sparsity to Entropy}

The direct way of utilizing these high zero proportions would be in form of a joint sparse \& quantized representation \citep{kurtic2022optimal, yu2023boost}: storing only the quantized values of non-zero weights, together with necessary position metadata. However, as our base quantization levels are already very low, standard sparsity metadata formats \citep{elsen2020fast, lin2023efficient} would only allow  limited additional compression. A bitmask indicating non-zero locations requires 1 bit per weight, while 10-13 bit (depending on layer size) column indices are even less memory efficient at the sparsity levels we encounter. Therefore, we take a different approach: we do not utilize sparsity directly but rather the \textit{low entropy}, which is implied by the fact that a single value (0) occurs very frequently, using an appropriate encoding scheme.

\subsubsection{Fast GPU Decoding Challenges}
\label{sec:decoding-challenges}

In principle, we could group multiple consecutive ternary weights into super-symbols and then apply a code which assigns \textit{variable length codewords} to those super-symbols, based on their probability of occurrence, for example, via a Huffman approach \citep{huffman1952method}. If the quantized weight values were close to independent, this would achieve strong compression rates; in fact, for actual independence, they would be essentially Shannon-optimal \citep{mackay2003information}.

At the same time, our primary goal is to use compressed models for \textit{fast and space-efficient inference}. Thus, it is critical not only that our encoding scheme achieves good compression, but also that it can be decoded fast on GPU hardware. This is challenging for a number of reasons:

\paragraph{Challenge 1:} Entropy-based codes generally possess sequential decoding dependencies: symbol $i$ can only be determined if the length, which is variable, of all ($i - 1$) prior symbols is known. Hence, processing consecutive symbols simultaneously leads to high synchronization overhead.

\paragraph{Challenge 2:} Binary words in storage (e.g., INT32 blobs) may contain different numbers of decoded symbols. Consequently, even if rows/blocks are encoded independently, parallel decoding will happen non-uniformly, while all threads in a GPU-warp must always execute the same instruction. This would result in many wasted operations.

\paragraph{Challenge 3:} Variable-length low-bit decoding involves a large number of binary operations like shifts, which are not particularly efficient on GPUs.

\paragraph{Challenge 4:} Individual matrices of MoEs are typically not very large, making it difficult to split them into enough separately decoded segments to achieve good GPU utilization without having to store additional data to break sequential dependencies, which would harm compression rates.

In contrast, uncompressed half-precision matrix-vector products, which are the primary operation underlying generative inference, easily achieve close to ideal memory-bandwidth utilization and thus present a very strong baseline.

\subsection{Compression Scheme \& Kernel Co-design}

To achieve our goal, we need to design a compression scheme and its GPU decoding kernel \textit{jointly}, and potentially trade off compression for faster decoding. We begin with an overview of the main ideas behind our approach, followed by an in-depth discussion of key details.

\subsubsection{Overview}

Instead of a code with variable length codewords (see Section~\ref{sec:decoding-challenges}) mapping to fixed length data, we will use a \textit{dictionary-based} code with fixed length codewords mapping to a variable number of symbols. Such LZW-based schemes \citep{welch1984technique} are popular for general purpose compression like ZIP, as they are particularly effective for text data with long repeated segments. While a dictionary code is not ideal in terms of compression rate for the case of almost-random data in our application, it will be key for fast GPU decoding.

First, our kernel design uses one warp, that is 32 consecutive threads, to handle a row of a weight matrix, each of which is encoded independently. This addresses Challenge 4 in Section~\ref{sec:decoding-challenges}, yielding reasonable GPU utilization for relevant matrix sizes, with negligible metadata overhead. Further, we use a fixed-to-variable code with a large dictionary. This allows us to use a full warp to process one codeword at-a-time, extracting all data, while maintaining good efficiency, thus working around Challenges 1 and 2. This way, slow bit and base-3 operations (for ternary) can also be kept at a minimum, resolving Challenge 3.

\subsubsection{Dictionary Design and Implementation}
\label{sec:dictionary}

In general, assume that the values of a ternary weight matrix (denoted by 0, 1, 2) are distributed close to independently according to the distribution:
\begin{equation}
    \label{eq:dist}
    P(0) = p_0, \quad P(1) = P(2) = \frac{1 - p_0}{2},
\end{equation}
where $p_0$ denotes the probability of sampling 0, e.g., 0.885 as per Table~\ref{tab:sparsity}. Since we plan to use a rather large dictionary, it should be shared between many weight matrices, in order for the dictionary itself not to cause substantial storage overheads. We find that such a static dictionary works well enough, while simplifying memory efficient compression (see Section \ref{sec:system-optimizations}) as we do not have to collect statistics over many yet uncompressed experts.

Next, we consider pairs of ternary values $t = (t_1, t_2)$, whose corresponding probability is $P(t) = P(t_1)P(t_2)$. We generate the $2^{16}$ highest probability sequences containing at most 14 such pairs. This dictionary can be generated using a max-priority queue on probability, as shown by Algorithm~\ref{alg:gen-dict}.

\begin{algorithm}[!ht]
    \centering
    \caption{Generate decoding dictionary sequences.}
    \label{alg:gen-dict}
    \begin{algorithmic}
        \STATE $Q \gets$ max priority queue containing $(1.0, ())$
        \WHILE{$|D| < 2^{16}$}
            \STATE $p, s \gets \text{pop}(Q)$
            \STATE append $s$ to dictionary if $0 < |s| < 28$
            \FOR{$t \in \{(t_1, t_2) \, | \, t_1, t_2 \in \{0, 1, 2\}\}$}
                \STATE $\text{push}((p \cdot P(t), \text{cat}(s, t)), Q)$
            \ENDFOR
        \ENDWHILE
    \end{algorithmic}
\end{algorithm}

To briefly understand the procedure, notice that upon the first iteration, it will push all individual pairs $t = (t_1, t_2)$ to the priority queue, sorting them by decreasing probability, after which they will be expanded in this order.

\begin{figure}[ht!]
    \centering
    \includegraphics[width=.95\linewidth]{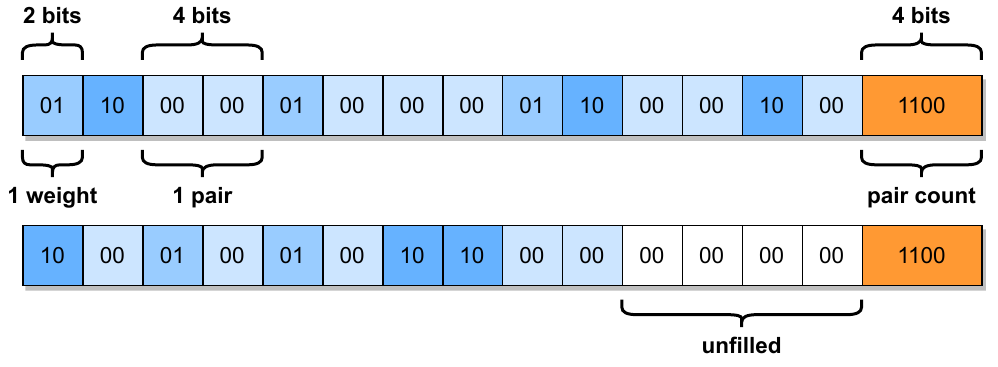}
    \vspace{-5pt}
    \caption{Data format of a dictionary entry; here of 24 weights.}
    \label{fig:encoding}
\end{figure}

We have exactly $2^{16}$ codewords as this allows us to store them in the native UINT16 datatype, avoiding any slow bit-extractions at this decoding level. Each of those codewords maps to two consecutive UINT32 values containing up to 7 pairs each, stored using 2 bits per ternary value, followed by the total number of pairs in the sequence; see also Figure~\ref{fig:encoding}. This format dictates our maximum chosen pair count of 14. Further, we consider pairs, rather than individual weights, to fit the maximum count into 4 bits. The 2-bit-per-weight format is used as there is enough space, while a more compact ternary encoding would involve slow modulo and division operations for extraction. We store the pair-count twice so that each thread can work with only half of the data, stored in a fast INT32 type.

Overall, mapping 16-bit codewords to 64-bit data blobs strikes a good balance between several goals: (a) Having codewords map to, on average, more uncompressed values than their bitwidth, a necessary condition for achieving $< 1$-bit compression. (b) Minimizing the overall storage cost of the dictionary to fit into the L2-cache of the GPU, which is critical for good decoding performance. (c) Utilizing as many threads in a warp as possible for simultaneously extracting plain weights from the decoded data; usually, $> 16$ will do useful work and only 4 out of 32 threads are never active in this step. (d) Avoiding as many conditionals and extra operations necessary for dealing with non-uniform data storage as possible, which slow down parallelization.

Finally, we note that while dictionary lookups are in principle random access, keeping it sorted from highest to lowest probability ensures very favorable caching behavior. Since each lookup also automatically prefetches several subsequent elements, and most lookups are for frequently occurring codewords, there are many fast L1-cache hits.

\paragraph{Validation.} To assess the effectiveness of our scheme, we compute achieved compression rates, both on a real ternary quantized c2048 model as well as on weight matrices sampled directly from distribution~(\ref{eq:dist}), yielding $20.07\times$ and $21.11\times$, respectively. This gap of only $\approx 5\%$ suggests that our simplifying independence assumption is indeed quite close for large models. We also note that our rates are only $\approx 20\%$ away from the distribution's (with $p = 0.885$) \textit{theoretical} compression limit of $25.40\times$, which we consider a reasonable trade-off for enabling fast GPU decoding.

\subsubsection{GPU Kernel}

Having defined the dictionary format, we can now discuss the design of the actual decoding kernel in detail. We focus on the most important operation for inference, decompression fused with a matrix-vector-product. However,  our techniques can easily be adapted to other use-cases, e.g., pure decompression.

Listing~\ref{lst:kernel} provides CUDA-like pseudocode for our kernel, computing the matrix-vector-product of compressed matrix \code{w\_comp} (with metadata \code{row\_off} and \code{ter\_minmax}, using dictionary \code{dec}) and BF16 vector \code{x}, into output buffer \code{y}. The handling of various edge cases and some index calculations have been removed for readability. Please see our repository for the fully functional implementation.

\definecolor{codegreen}{rgb}{0,0.6,0}
\definecolor{codegray}{rgb}{0.5,0.5,0.5}
\definecolor{codepurple}{rgb}{0.58,0,0.82}
\definecolor{backcolour}{rgb}{0.95,0.95,0.92}

\lstdefinestyle{codestyle}{
    backgroundcolor=\color{backcolour},   
    commentstyle=\color{codegreen},
    keywordstyle=\color{magenta},
    numberstyle=\tiny\color{codegray},
    stringstyle=\color{codepurple},
    basicstyle=\ttfamily\tiny,
    breakatwhitespace=false,         
    breaklines=true,                 
    captionpos=b,                    
    keepspaces=true,                 
    numbers=left,                    
    numbersep=5pt,                  
    showspaces=false,                
    showstringspaces=false,
    showtabs=false,
    tabsize=1,
}

\lstset{style=codestyle}

\begin{lstlisting}[
    label={lst:kernel},
    caption={Simplified kernel pseudocode for a fused decompress + matrix-vector-product operation.}
]
template <int num_warps, int w_width>
__global__ void Sub1MatVec(
  int* dec,
  ushort* w_comp, int* row_off, __nv_bfloat162* ter_minmax,
  __nv_bfloat16* x, __nv_bfloat16* y
) {
  __shared__ float x_shared[w_width];
  for (int i = thread; i < w_width; i += 32 * num_warps)
    x_shared[i] = __bfloat162float(x[i]);
    
  __shared__ float deq[3][32 * num_warps];
  deq[0][thread] = 0;
  deq[1][thread] = __bfloat162float(ter_minmax[row].x);
  deq[2][thread] = __bfloat162float(ter_minmax[row].y);

  __syncthreads();
  __shared__ w_comp_block[32][num_warps];
  
  float res = 0;
  int idx = 0;

  for (int i = 0; i < row_off[row + 1] - row_off[row]; i += 32) {
    w_comp_block[warp][lane] = w_comp[i + lane];

    if (lane < 28) {
      for (int j = 0; j < 32; j++) {
        int enc = w_comp_block[warp][j];
        int wx14 = dec[2 * enc + (lane / 14)];
        int ter = (wx14 >> (4 + 2 * (lane % 14))) & 0x3;
        float w = deq[ter][thread];
        res += w * x_shared[idx + lane];
        idx += 2 * (wx14 & 0xf);
      }
    }
  }

  for (int i = 16; i > 0; i /= 2)
    res += __shfl_down_sync(0xffffffff, res, i);
  if (lane == 0)
    y[row] += __float2bfloat16(res);
}
\end{lstlisting}

\paragraph{Parallelization.} Overall, each threadblock will handle multiple consecutive rows, each of which is processed by a single warp. We use exactly one thread-block per GPU Streaming Multiprocessor (SM) with $\text{min}(\text{\#rows\_in\_block}, 32)$ warps; if there are more than 32 rows in a block, (some) warps sequentially process multiple rows (note that this part is omitted in Listing~\ref{lst:kernel} for simplicity). This avoids any bad wave quantization effects. We find this strategy to be an effective heuristic that yields good performance for all matrix shapes we consider.

\paragraph{Execution.} Our kernel starts by loading the entire input vector to shared memory (\code{x\_shared}, lines 7-9), using all warps in a threadblock. This enables fast element access in the subsequent per-row product-sum accumulations.

Next, each warp processes its corresponding row by first fetching (up to) 32 codewords into shared memory (\code{w\_comp\_block}, line 23) using a single coalesced transaction. It then loops over those symbols, processing one-at-a-time (lines 26-33). First, using 28 of its 32 threads (line 25), it fetches the corresponding decoding data from the dictionary where the first UINT32 is assigned to threads 0-13 and the second to threads 14-27 (\code{wx14}, line 27). Then, each thread extracts its corresponding ternary weight (lines 29-30) and adds the corresponding input product into its own partial result accumulator (\code{res}, line 31). We note that the input reads from shared memory are contiguous and do not cause bank conflicts. Afterwards, each thread advances the offset index (\code{idx}, line 32) into the input vector by the total number of weights encoded in the current symbol.

Finally, after the full row has been scanned, a warp-reduction (lines 37-38) over the partial results of each thread yields the output (\code{y}, lines 39-40).

\paragraph{Ternary decoding.} Another relevant detail is that ternary weights are stored as $0,1,2$ (line 29) but need to be dequantized to $0,w_\text{min}, w_\text{max}$ for multiplication with inputs. We found that the most efficient way of performing this conversion is via a shared memory lookup table (lines 11-14). Crucially, this table needs to be replicated 32 times across the column-dimension to avoid very frequent bank conflicts, which would otherwise occur every time not all 28 threads dequantize the same value (line 30). Fortunately, there are only 3 input values and so its overall size is tolerable.

\paragraph{Encoding.} So far, we have only focused on the decoding operation, but we also have to \textit{encode} matrices with reasonable efficiency. In general, this is done by building a trie datastructure (of the dictionary discussed in Section~\ref{sec:dictionary}) mapping sequences to codewords. Then, we iterate through the input while simulatenously traversing the trie to find longest prefix matches, yielding the corresponding codewords. Finally, we densely pack rows of different lengths into a contiguous buffer and record corresponding row offsets. Unlike decoding, encoding is not very latency critical and a straight-forward GPU kernel using one thread per row of the matrix to compress suffices.

\section{Experiments}

\subsection{General Setup}
\label{sec:experiment-setup}

\paragraph{Models.} We focus our experiments on the SwitchTransformer~\citep{fedus2022switch} family of models. Our primary target is the very largest variant, c2048, with around 1.6 trillion parameters, but we also consider the comparatively small base128 (7B params) and large128 (26B params) versions for testing and ablations. We chose the SwitchTransformer family as it contains the largest publicly-available model, which also features a similar or higher number of training tokens to parameters ratio than potential alternatives like \citet{artetxe2021efficient}. Further, those models are also among the most popular massive MoEs, with several implementations across frameworks.

\paragraph{Framework.} As accessibility is a major goal of our work, we build our code-base around the PyTorch-backend of the highly popular HuggingFace~\citep{wolf2019huggingface} framework, rather than on the SwitchTransormer's original training environment MeshTensorflow~\citep{shazeer2018mesh} or its JAX-based successor T5X~\citep{t5-code}. This brings a number of additional challenges.

First, we find that the largest model variants require a handful of bugfixes, primarily configuration and model setup changes, in order to run properly. We suspect that this is because their enormous sizes have rendered extensive testing very difficult. Second, we observed a major inefficiency in the context of generative inference for models with a large number of experts: the HuggingFace implementation will perform several (empty) CUDA calls for potentially 1000s of experts to which no token is routed, accumulating large overheads. We modify the implementation (also for baselines) to skip such unnecessary calls, leading to $> 10\times$ speedup for large models. We apply all changes to the HuggingFace framework only dynamically at runtime, so that our code can be run directly with an official installation.

HuggingFace prioritizes ease-of-use and flexibility over high performance. For that reason, we conduct inference measurements not only end-to-end, including all HuggingFace overheads, but also in isolated fashion, comparing uncompressed and compressed matrix operations directly. This is to demonstrate that our GPU kernels would also yield low overhead in more optimized inference environments.

\paragraph{Datasets.} SwitchTransformers have been trained for a Masked-Language-Modelling (MLM) objective~\citep{raffel2020exploring} on the C4 dataset~\citep{C4}. Similar to most works in the area of LLM quantization \citep{yao2022zeroquant, frantar2022gptq, dettmers2022case}, we focus on general \textit{upstream} compression directly on this pretraining task/dataset combination. Consequently, our evaluation focuses on validation performance for C4/MLM, where we use the public reproduction of C4 on HuggingFace as well as their replication of the original masking procedure. Calibration data for compression is taken, in order, from the first two shards of the training set. For efficiency, we primarily evaluate on 128 samples (corresponding to the average loss over $>$ 10K tokens, which is quite stable) from the first shard of the validation set, but we also perform some evaluations other datasets.

\paragraph{Hardware.} All compression experiments, including those for the very largest models, can be performed in less than a day on a single NVIDIA A6000 with 48GB of GPU memory. However, efficiently compressing trillion parameter models using a large number of calibration samples requires a few 100GBs of (CPU) RAM; the original 1.6T model itself also occupies $> 3$ TB disk storage. We highlight that our work is performed in a highly constrained environment for models of this size, for example, it is already infeasible to load the entire (uncompressed) 1.6T model into RAM, let alone into GPU memory. For inference on compressed models, we will also consider running on multiple NVIDIA 3090 GPUs, with 24GB of memory each, in addition to A6000s.

\subsection{Compression Results}
\label{sec:compression-results}

\paragraph{Accuracy.} We begin by quantizing all SwitchTransformer models to 2-bit and ternary precision, and evaluating their validation loss. Our default number of calibration samples is 10K for 128 experts and 160K for 2048, but we also consider using $0.5\times$ and $2\times$ as many samples. In addition to using our efficient QMoE framework discussed in Section~\ref{sec:scaling-quantization}, we also consider a standard round-to-nearest (RTN) baseline \citep{dettmers2022llm}. We simulate the latter by fixing Hessians to the identity matrix, thus applying precisely the same quantization settings and evaluation protocol. Table~\ref{tab:accuracy-results} summarizes our results.

\begin{table}[ht!]
    \centering
    \begin{tabular}{|c|cc|cc|cc|}
        \toprule
        \multirow{2}{*}{method} & \multicolumn{2}{c|}{base128} & \multicolumn{2}{c|}{large128} & \multicolumn{2}{c|}{c2048} \\
         &  2bit & tern & 2bit & tern & 2bit & tern \\
        \midrule
        BF16 & \multicolumn{2}{c|}{1.73} & \multicolumn{2}{c|}{1.55} & \multicolumn{2}{c|}{1.18} \\
        \midrule
        RTN & 2.27 & 4.54 & 1.96 & 2.79 & 1.33 & 2.15 \\
        \midrule
        QMoE 0.5x & 1.78 & 2.11 & 1.54 & 1.70 & 1.22 & 1.27 \\
        QMoE 1.0x & \textbf{1.76} & 1.99 & \textbf{1.56} & 1.69 & \textbf{1.20} & \textbf{1.26} \\
        QMoE 2.0x & \textbf{1.76} & \textbf{1.93} & 1.57 & \textbf{1.64} & 1.21 & \textbf{1.26} \\
        \bottomrule
    \end{tabular}
    \vspace{-5pt}
    \caption{Comparing C4 validation losses for 2-bit and ternary (tern) quantized SwitchTransformers. ``QMoE 0.5x'' indicates that only half of the default number of calibration samples are used. }
    \label{tab:accuracy-results}
\end{table}

Perhaps surprisingly, vanilla rounding (RTN) does not lead to a complete model collapse even at ternary precision, emphasizing the high robustness of large MoEs to quantization. Nevertheless, the loss increases are quite significant for smaller models at 2-bit and far too large to be useful at ternary precision. In contrast, using data-dependent quantization, 2-bit is achievable at minimal loss (1.7\% relative on c2048) and ternary at only a small increase (6.7\% relative on c2048). This demonstrates not only the effectiveness of such advanced quantization methods in this context, but also shows that extremely low-bit compression is indeed practical for massive MoEs.

Additionally, we conduct evaluations on Arxiv, GitHub, StackeEchange and Wikipedia data sampled from RedPajama \citep{together2023redpajama}. Even though only $< 0.01$\% of our C4 calibration data originates from those websites, the compressed model still preserves performance almost as well as on the core of the distribution (see Table~\ref{tab:extra-evals}).

\begin{table}[ht!]
    \centering
    \begin{tabular}{|c|cccc|}
        \toprule
        bits & arxiv & github & stackexch. & wiki \\
        \midrule
        BF16 & 1.31 & 0.99 & 1.15 & 1.20 \\
        \midrule
        2-bit & 1.34 & 1.05 & 1.17 & 1.24 \\
        tern & 1.42 & 1.13 & 1.22 & 1.32 \\
        \bottomrule
    \end{tabular}
    \vspace{-5pt}
    \caption{Additional evaluations for the c2048 model.}
    \label{tab:extra-evals}
\end{table}

In terms of calibration data, we see that increasing the amount of samples generally improves performance slightly, most noticeably for ternary quantization, but there is also some noise in the process, especially at 2-bit.

\paragraph{Compression.} Next, we investigate the actual compression rates that are achieved by further compressing ternary models using our scheme introduced in Section~\ref{sec:realizing-sub1}. We consider both compression relative to just the MoE modules (the model parts we quantize) as well as to the full model and all its metadata. The compression rates and overall checkpoint sizes are listed in Table~\ref{tab:compression-rates}.

\begin{table}[ht!]
    \centering
    \begin{tabular}{|c|cc|cc|}
    \toprule
    \multirow{2}{*}{model} & \multirow{2}{*}{moe-only} & \multirow{2}{*}{full} & \multicolumn{2}{c|}{size [GB]} \\
    & & & bf16 & ours \\
    \midrule
    base128 & $17.06\times$ & $11.76\times$ & 14.9 & 1.27 \\
    large128 & $18.34\times$ & $13.32\times$ & 52.7 & 3.96 \\
    c2048 & $20.07\times$ & $19.81\times$ & 3142 & 158.6 \\
    \bottomrule
    \end{tabular}
    \vspace{-5pt}
    \caption{Compression rates and sizes for ternary models.}
    \label{tab:compression-rates}
\end{table}

In general, measuring only relative to parts we compress (moe-only), all sizes achieve $> 16\times$ compression rate and thus $< 1$ bits per parameter storage. On c2048, even the overall rate, including all uncompressed dense layers, remains at $19.81\times$, corresponding to \textit{0.807 bits per parameter}, reducing the checkpoint size from 3142GB to 158.6GB. One can also observe that compression rates increase with model size, which is for two reasons: (a) natural sparsity increases while our encoding dictionary is also optimized for c2048 (see Section~\ref{sec:realizing-sub1}), and (b) weight distributions become closer to independent for larger layer sizes.

\begin{figure*}[t]
    \centering
    \begin{subfigure}{.5\textwidth}
      \centering
      \includegraphics[width=.9\linewidth]{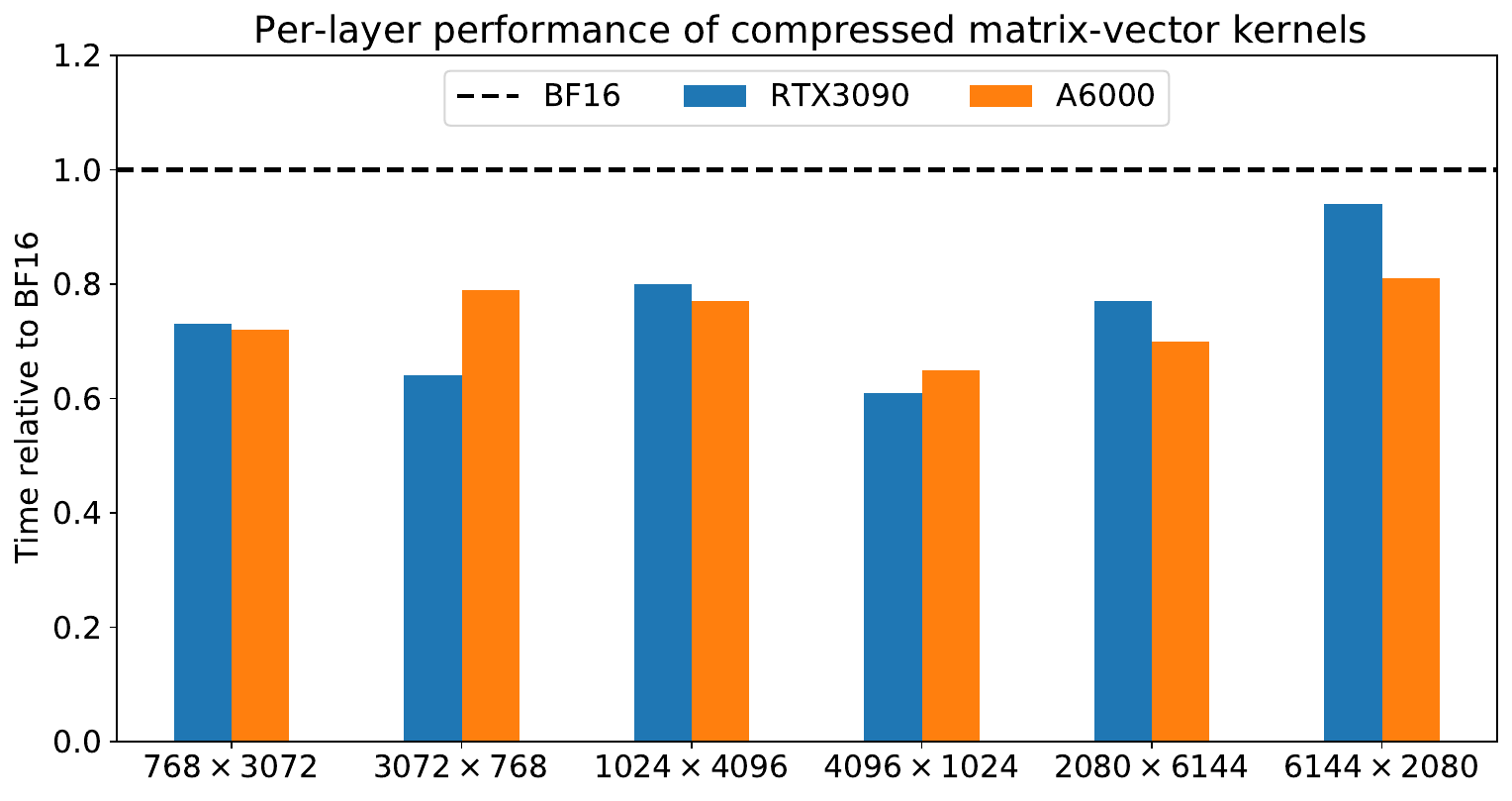}
    \end{subfigure}%
    \begin{subfigure}{.5\textwidth}
      \centering
      \includegraphics[width=.9\linewidth]{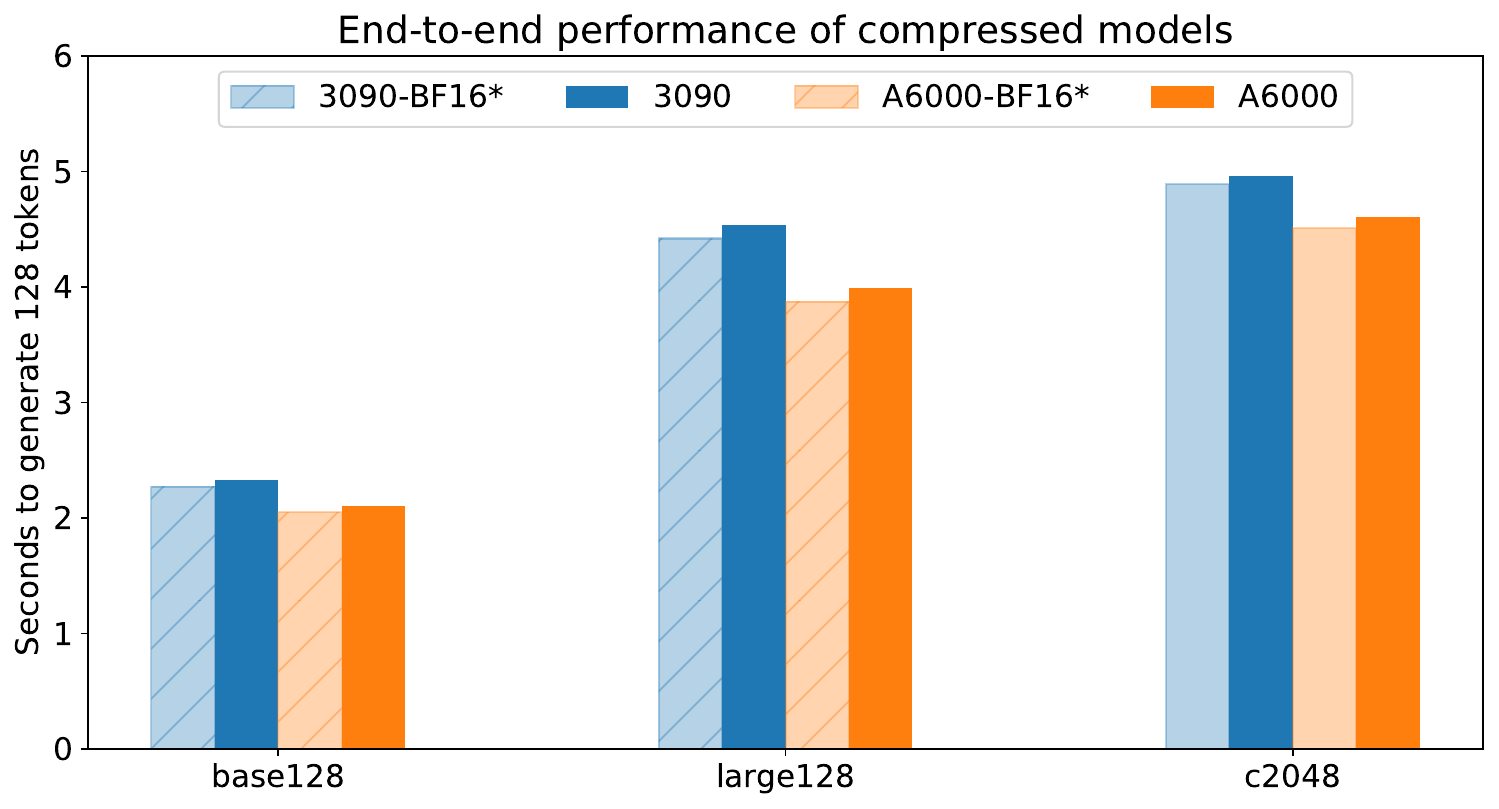}
    \end{subfigure}
    \vspace{-5pt}
    \caption{(Left) Per-layer compressed kernel performance relative to uncompressed execution. (Right) End-to-end runtimes of compressed models and estimates ($^*$, would require 65/130 GPUs) for bloat16 baselines. c2048 is run on 4$\times$A6000 and 8$\times$3090 GPUs, respectively.}
    \label{fig:kernel-performance}
\end{figure*}

\paragraph{Runtime.} Finally, we evaluate how long it takes to produce compressed models on a single A6000 GPU, for different amounts of calibration data. The results are shown in Table~\ref{tab:runtimes}. Smaller models can be compressed in less than an hour and even c2048 in less than a day, confirming the high efficiency of QMoE. The runtime increase from large128 to c2048 is roughly proportional to the difference in size, despite the latter using $16\times$ more samples. This is because the number of samples per expert stays constant and the expert size increases only slightly. Finally, we note that simply (iteratively) loading the original 1.6T model into RAM takes close to 5 hours on our slow disk storage.

\begin{table}[ht!]
    \centering
    \begin{tabular}{|c|ccc|}
        \toprule
        model & 5K/80K & 10K/160K & 20K/320K \\
        \midrule
        base128 & 8.4min & 14.0min & 21.6min \\
        large128 & 22.0min & 30.2min & 45.2min \\
        c2048 & 13.3h & 16.0h & 20.8h \\
        \bottomrule
    \end{tabular}
    \vspace{-5pt}
    \caption{Compression runtime for different calibration data size.}
    \label{tab:runtimes}
\end{table}

\subsection{Runtime Results}

\paragraph{Individual Layers.} Our kernel performance evaluation starts with a direct (isolated) comparison of our compressed matrix-vector product kernels (see Section~\ref{sec:realizing-sub1}) against PyTorch's standard (uncompressed) bfloat16 cuBLAS kernels. Figure~\ref{fig:kernel-performance} (Left) shows the time taken by our compressed kernels relative to bfloat16, for the matrix shapes found in our MoEs, on two different GPUs. While our kernels have to perform a lot less slow (global) memory reads than the bfloat16 baseline due to lower storage costs, they need to spend much more compute for complex unpacking of the heavily-compressed weights. Nevertheless, executing our compressed kernels takes less time than the close to ideal bfloat16 baseline in all cases, with up to 35\% speedup on specific matrix shapes. We note that these are very low-latency operations, with the smallest matrix taking $< 0.02$ milliseconds and the largest $< 0.05$.

\paragraph{End-to-End Execution.} Finally, we also benchmark our kernels end-to-end in HuggingFace on the real weights of our compressed MoE models. We consider an individual user application, like \citep{frantar2022gptq, leviathan2023fast, park2022nuqmm}, where a single prompt (sampled from C4) should be processed to generate a 128-token response. As actually running the bfloat16 version of the c2048 model would require $> 65$ A6000 and $> 130$ 3090 GPUs (versus 4 and 8, respectively, for sub-1-bit compressed weights) we have to estimate its runtime. We do this by having all experts in a layer point to the same weight data (completely resolving memory issues), which allows us to collect timings with precisely the same overheads as for our compressed models. However, this is a highly optimistic estimate since real execution would require close to $20\times$ more GPUs, with corresponding communication overheads, and our numbers should thus be viewed only as a lower bound.

The results, shown in Figure~\ref{fig:kernel-performance} (Right), demonstrate that end-to-end execution of compressed models is only $< 5\%$ slower than standard (uncompressed) execution. This slight slow-down despite faster per-layer timings is due to the fact that the encoder may sometimes route multiple tokens to the same expert. Our current implementation naively executes a separate matrix-vector product for each token, while the baseline performs a much more efficient joint matrix multiplication. For applications where this is a significant bottleneck, one could easily introduce an inner loop over tokens into our kernel (Listing~\ref{lst:kernel}, line 30), or fully decompress first, followed by a standard matmul, for large token counts.

\section{Related Work}

\paragraph{Mixture-of-Expert (MoE) Models.} Mixture-of-expert models are a popular research direction aimed at creating significantly more efficient large-scale models \citep{fedus2022switch, artetxe2021efficient, clark2022unified}. At the core of MoEs lie (sparse) routing mechanisms, of which many variants have been proposed. Those range from static assignment based on input tokens IDs \citep{roller2021hash}, over dynamic token-to-expert matching \citep{zhou2022mixture}, to ``soft'' routing of linear input combinations \citep{puigcerver2023sparse}. Since MoEs can feature rather different computational profiles from standard dense models, there is also significant research on optimizing inference and training systems \citep{barham2022pathways, gale2023megablocks, hwang2023tutel}. Among the most critical problems in this area are data-exchanges between accelerators during routing and dealing with uneven compute-loads for different experts.

\paragraph{LLM Quantization.} Quantization is a very popular compression technique, which has seen a vast amount of work~\cite{gholami2021survey}, especially in the context of LLMs. Specifically, the ability to perform accurate weight quantization for billion-parameter models has greatly boosted their accessibility: it has been shown that extremely large dense models can be quantized to 8- or even 4-bit precision at little accuracy loss \citep{dettmers2022llm, yao2022zeroquant, frantar2022gptq, dettmers2022case}. 
Pushing towards even lower bitwidths via more sophisticated compression formats, like multi-level grouping coupled with higher-precision outliers \citep{dettmers2023spqr}, or new quantization techniques, like incoherence preprocessing~\citep{chee2023quip}, is an active area of research.
Currently, accurate quantization to 2 or less bits per parameter appears to be a major barrier for post-training quantization of standard LLMs.
By contrast, in this work we show that massive MoE models appear to be significantly more compressible, as we achieve sub-1-bit compression at comparable loss increases to 3-bit or 4-bit quantization of standard LLMs with advanced techniques.

\paragraph{MoE Compression.} There has also been work on compressing MoE models in particular. \citet{chen2022task} and \citet{koishekenov2022memory} perform compression via specialization of MoEs to specific ``downstream'' finetuning datasets by pruning components not relevant to the particular task.
In contrast, we focus on general ``upstream'' compression of the pretrained model, via extremely low-bit quantization. Other works \citep{kim2022says, yi2023edgemoe, kim2023finequant} also perform MoE quantization, but focus on noticeably higher bit-widths, like 8 or 4 bits per weight. This is accomplished primarily via simple rounding, which, as shown by our experiments, is not accurate enough for full 2-bit or lower compression.
\citet{kim2022mixture} achieve 2-bit quantization on a 5 billion parameter MoE, which is considered relatively small in this area, by further optimization of the model via Quantization-Aware Training \citep{nagel2021white}. 
Applying such an approach for trillion-scale models would be extremely resource intensive. They also do not provide any mechansims for exploiting low-bit quantization and its corresponding natural sparsity in practice, which is challenging and constitutes a key contribution of our work.

We are particularly focused on scalabilty and practicalty. While existing works study models with at most tens of billions of parameters, we demonstrate the effectiveness and efficiency of our techniques at trillion parameter scale, both for the quantization process itself as well as for actual inference of compressed models.

\section{Discussion and Limitations}

We have presented QMoE, an end-to-end compression and inference framework for addressing the massive memory costs of MoE inference. We showed, for the first time, that models such as the trillion-parameter SwitchTransformer-c2048 can be accurately compressed to less than 1 bit per parameter, close to $20\times$ compression rate, in a custom format that enables the first efficient end-to-end execution of such a model on a single commodity GPU server. QMoE is fully open-source and built around the popular HuggingFace framework, making deployment and research for massive MoEs significantly cheaper and more accessible.

Our study is confined to a limited set of models, as only very few massive and accurate MoEs are available publicy. Additionaly, due to their size, most MoEs are trained and deployed in different bespoke framework, requiring complex manual integrations to use for further research. Nevertheless, we have covered some of the largest and most accurate available MoEs, specifically SwitchTransformers~\cite{fedus2022switch}. A natural extension of our work would be to apply our QMoE techniques to other MoE models and variants, such as~\citet{artetxe2021efficient} or the recently-proposed SoftMoEs~\cite{puigcerver2023sparse}.

Additionally, we have focused on direct compression of the pretrained base model. However, it would also be interesting to further finetune a compressed model for specialized down-stream tasks, similar to QLoRA~\citep{dettmers2023qlora}. \citet{zoph2022st} report strong results when finetuning only non-expert layers, which QMoE leaves uncompressed, suggesting that this application could be promising. We hope to explore this in future work.

\bibliography{references}
\bibliographystyle{arxiv}

\newpage
\appendix
\onecolumn

\end{document}